\documentclass{article}

\usepackage{PRIMEarxiv}

\usepackage[utf8]{inputenc} 
\usepackage[T1]{fontenc}    
\usepackage{hyperref}       
\usepackage{url}            
\usepackage{booktabs}       
\usepackage{amsfonts}       
\usepackage{nicefrac}       
\usepackage{microtype}      
\usepackage{lipsum}
\usepackage{fancyhdr}       
\usepackage{graphicx}       
\graphicspath{{media/}}     
\usepackage[numbers,sort&compress]{natbib}
\title{AI-Generated Content Enhanced Computer-Aided Diagnosis Model for Thyroid Nodules: A ChatGPT-Style Assistant
}

\author{
Jincao Yao$^{1,2,3,4,5,6\#}$, Yunpeng Wang$^{7\#}$, Zhikai Lei$^{8\#}$, Kai Wang$^{9\#}$, Xiaoxian Li$^{10}$, Jianhua Zhou$^{10}$, \\
{\bf Xiang Hao$^{7}$, Jiafei Shen$^{1,2}$, Zhenping Wang$^{9}$, Rongrong Ru$^{11}$, Yaqing Chen$^{11}$, Yahan Zhou$^{6}$,}\\
	{\bf Chen Chen$^{1,2}$,  Yanming Zhang$^{12,13*}$, Ping Liang$^{14*}$, Dong Xu$^{1,2,3,4,5,6*}$} \\
$^{1}$Department of Radiology, Zhejiang Cancer Hospital, Hangzhou, 310022, China\\
$^{2}$Hangzhou Institute of Medicine (HIM), Chinese Academy of Sciences, Hangzhou, 310000, China\\
$^{3}$Key Laboratory of Head \& Neck Cancer Translational Research of Zhejiang Province,\\ Hangzhou, 310022, China\\
$^{4}$Zhejiang Provincial Research Center for Cancer Intelligent Diagnosis and Molecular Technology,\\ Hangzhou, 310000, China\\
$^{5}$Wenling Medical Big Data and Artificial Intelligence Research Institute, 24th Floor, Machang Road,\\ Taizhou, 310061, China\\
$^{6}$Taizhou Key Laboratory of Minimally Invasive Interventional Therapy \& Artificial Intelligence,\\ Taizhou Campus of Zhejiang Cancer Hospital (Taizhou Cancer Hospital), Taizhou, 317502, China\\
$^{7}$College of Optical Science and Engineering, Zhejiang University,\\ No.38 of Zheda Road, Hangzhou, Zhejiang Province, China\\
$^{8}$Zhejiang Provincial Hospital of Chinese Medicine, 54 Youdian Road, Hangzhou, 310003, China\\
$^{9}$Department of Ultrasound, The Affiliated Dongyang Hospital of Wenzhou Medical University,\\ Dongyang, 322100, China\\
$^{10}$Department of Ultrasound, Sun Yat-sen University Cancer Center, State Key Laboratory of Oncology\\ in South China, Collaborative Innovation Center for Cancer Medicine, Guangzhou, 510060, China\\
$^{11}$Affiliated Xiaoshan Hospital, Hangzhou Normal University, No.728 North Yucai Road,\\ Hangzhou, 311202, China\\
$^{12}$Zhejiang Provincial People's Hospital Affiliated People's Hospital, \\Hangzhou Medical College, Hangzhou, 314408, China\\
$^{13}$Key Laboratory of Endocrine Gland Diseases of Zhejiang Province, Hangzhou, 314408, China\\
$^{14}$Department of Ultrasound, Chinese PLA General Hospital, Chinese PLA Medical School,\\ Beijing, 100853, China
}

\begin{document}
\maketitle

\begin{abstract}
An artificial intelligence-generated content-enhanced computer-aided diagnosis (AIGC-CAD) model, designated as ThyGPT, has been developed. This model, inspired by the architecture of ChatGPT, could assist radiologists in assessing the risk of thyroid nodules through semantic-level human-machine interaction. A dataset comprising 19,165 thyroid nodule ultrasound cases from Zhejiang Cancer Hospital was assembled to facilitate the training and validation of the model. After training, ThyGPT could automatically evaluate thyroid nodule and engage in effective communication with physicians through human-computer interaction. The performance of ThyGPT was rigorously quantified using established metrics such as the receiver operating characteristic (ROC) curve, area under the curve (AUC), sensitivity, and specificity. The empirical findings revealed that radiologists, when supplemented with ThyGPT, markedly surpassed the diagnostic acumen of their peers utilizing traditional methods as well as the performance of the model in isolation. These findings suggest that AIGC-CAD systems, exemplified by ThyGPT, hold the promise to fundamentally transform the diagnostic workflows of radiologists in forthcoming years.
\end{abstract}


\section{Introduction}
Thyroid nodules are a common endocrine disorder, with up to 68\% of adults afflicted, and approximately 7-15\% of these nodules are thyroid cancer\cite{1,2,3}. Ultrasound is the preferred imaging diagnostic method for assessing thyroid nodules due to its non-invasive, radiation-free, and user-friendly nature\cite{3}. However, ultrasound diagnosis is highly dependent on the clinician's experience, leading to subjectivity and inconsistencies\cite{4}. To provide a more objective and accurate assessment of thyroid nodules, many researchers have begun to build Computer-Aided Diagnosis (CAD) models. These models aim to extract features from ultrasound images to assist clinicians in a more objective and precise evaluation. A series of studies have shown that CAD models based on deep learning, such as ThyNet and RadImageNet\cite{5,6,7}, have achieved good results.

Despite these advancements, the aforementioned CAD methods have significant flaws that cannot be ignored. On one hand, traditional CAD models fail to provide the rationale behind their diagnoses or the decision-making process of their analysis, creating a gap in understanding between doctors and CAD models\cite{7,8}. This ``black box" characteristic undermines the confidence of doctors, patients, and healthcare administrators in the diagnostic results\cite{9}. On the other hand, most existing CAD models only mechanically provide pattern recognition probabilities, such as the probability of malignancy or metastasis, without further interaction with clinicians\cite{10,11}. This ``mute box" characteristic leads many clinicians to abandon these CAD models in favor of traditional diagnostic methods that are transparent, understandable, and interpretable\cite{12,13,14,15,16}. This phenomenon has sparked debate about the actual role of artificial intelligence in medical imaging applications and has hindered the development and application of big data and artificial intelligence technologies in the field of assisted diagnosis to some extent\cite{17,18,19,20}.

Recently, generative large language models (LLMs), represented by ChatGPT, have developed rapidly, demonstrating superior performance to other traditional models in semantic understanding, human-computer interaction, and robotics\cite{21,22,23,24,25}. The emergence of these LLMs provided an opportunity to bridge the ``interaction" and ``understanding" gaps between doctors and AI models\cite{26,27,28,29,30,31}. Meanwhile, it also lays the foundation for building the next generation of AI-generated content enhanced computer-aided diagnosis (AIGC-CAD) frameworks. In this study, we have constructed a large language model named Generative Pre-trained Transformer for Thyroid Nodules (ThyGPT), which focuses on the risk assessment of thyroid nodules. To train ThyGPT effectively, we retrospectively collected ultrasound images, anonymized ultrasound diagnostic reports, the thyroid nodule diagnostic guidelines and research reports from nited States, Europe et al., as training corpora. After training, ThyGPT is able to intuitively display the decision-making rationale of the CAD model and the contribution of ultrasound features of interest to clinicians in assisted diagnosis through human-computer interaction. Compared to previous generations of diagnostic models that could only output probabilities of benign or malignant conditions, the ThyGPT model makes the logic of AI analysis more transparent. It also allows clinicians to observe and consider various intermediate results during the model-assisted diagnostic process, significantly enhancing confidence in the use of CAD models.

The innovation of this study is twofold: (1) To our knowledge, ThyGPT is the first large language model study in the field of AI-assisted diagnosis of thyroid nodules. The model is the first to attempt to build a LLM for thyroid nodule risk using a large scale of comprehensive construction of multi-source information, including doctors' diagnostic reports, pathological results, international diagnostic guidelines, research reports, and ultrasound images. (2) We introduce the concept of AIGC-CAD for the first time in this study. AIGC-CAD enables the model to use generative large models to generate explanatory texts and feature markings for different components of the lesion and image semantics involved in the AI-assisted diagnostic process. This approach provides a more intuitive display and analysis, thereby better assisting clinicians. The ThyGPT constructed in this paper is only a preliminary exploration of this new paradigm in the thyroid field. This new CAD paradigm has the potential to become the mainstream direction for the next generation of CAD and to completely change the way clinicians use CAD tools.

\section{Materials and Methods}
\subsection{Patients}
This retrospective study was approved by the ethics committee of the participating hospital and was exempted from informed consent. The study was conducted in accordance with the ethical guidelines of the Declaration of Helsinki, and all data were anonymized. Initially, we randomly collected a total of 135,861 thyroid nodule ultrasound images, which came from 21,674 patients, all of whom received pathological results. We also collected 12,781 anonymized ultrasound diagnostic reports and over 1 million words of related text, including ACR and EU thyroid nodule diagnostic guidelines, to build the training and testing sets for the model.

Inclusion criteria for retrospective cases were: (1) patients found to have thyroid nodules during ultrasound examination and had ultrasound images recorded; (2) patients subsequently underwent total thyroidectomy or thyroid lobectomy and had pathological results recorded; (3) patients received fine-needle aspiration and obtained definitive positive results. Exclusion criteria were: (1) missing or incomplete corresponding ultrasound images of thyroid nodules, e.g., nodules too large to be displayed in a single image or key images missing; (2) patients received other types of treatment before surgery, such as I131 treatment; (3) incomplete clinical information of patients, such as missing basic patient information or previous treatment history. After exclusions, data from 19,165 patients and 12,073 anonymized ultrasound diagnostic reports were retained. Out of these, 15,136 patients were randomly selected as the training set, 1,933 patients as independent test set 1, and 2,096 patients as independent test set 2.

\subsection{Overall Design}
Figure 1 presents the overall design of our proposed AIGC-CAD model for the auxiliary diagnosis of thyroid nodules. Initially, we input pre-labeled nodule pathological information, anonymized diagnostic reports, thyroid-related diagnostic guidelines, research documents, and manually delineated nodule masks and bounding boxes into the model for training and parameter optimization of the ThyGPT model. Once training was completed, the relevant parameters of the ThyGPT model were fixed. Then, data from the independent test sets were input into the ThyGPT model for nodule analysis and auxiliary diagnosis. The model can automatically scan typical ultrasound features, such as calcification, margin conditions, etc., and combine feature heatmaps to analyze which features contribute most to the final differentiation.
\begin{figure}[htp]
	\centering
	\includegraphics[width=0.9\linewidth]{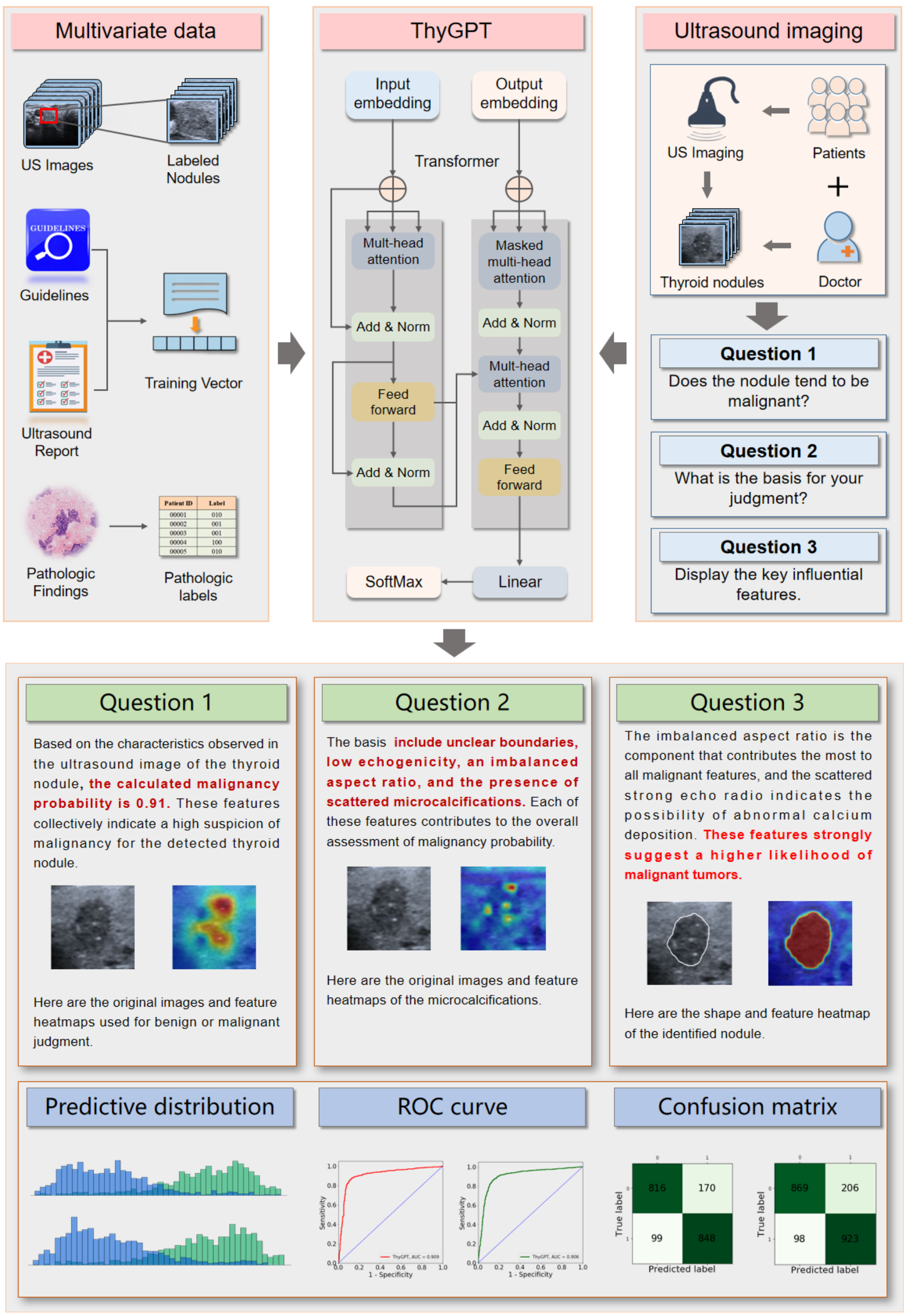}
	\caption{The overall design of our proposed ThyGPT model for thyroid nodules.}
	\label{fig:cad-01}
\end{figure}

\subsection{Model}
The basic framework used in this paper is the LlaMA2-13B model\cite{21}, which was further trained on this foundation, including supervised training involving instructions for tasks describing thyroid cases and the linguistic habits of thyroid doctors. Bleu scores and human evaluation were used as comprehensive assessment indicators. The backbone network of the model used the Lang-Chain framework, which is based on the GPT series design, and we utilized document loaders, text segmenters, and vector repositories for segmenting, vectorizing, and storing thyroid-related knowledge and cases. The image analysis model was trained on a hybrid of Swin-Transformer and DCNN models\cite{23,27}, which refer to self-attention, convolution and encoder-decoder architecture. Finally, we fused the image analysis model and the nodule recognition model to establish the ThyGPT model. Figure 2 shows the internal design architecture of our large model, including the backbone network, image analysis network, and nodule recognition network.

\subsection{Image Acquisition}
During image acquisition, patients were placed in the supine position with the neck extended and temporarily refrained from swallowing to fully expose the neck. The examining physician scanned the thyroid and stored at least one cross-sectional, longitudinal, or characteristic sectional ultrasound image. The ultrasound imaging data retrospectively collected in this study came from 31 machines, including GE, Siemens, Toshiba, and Philips. All radiologists collecting ultrasound images were professionally trained, and all data were quality-controlled by at least one senior ultrasound radiologist with over 10 years of experience.

\subsection{Annotation and Preprocessing}
In this study, we used mask annotation to depict the shape of nodules and areas of interest with different semantics; In addition, we recorded the benign or malignant nature of nodules, calcification, margins, echogenicity, etc. All ultrasound images were anonymized, removing any information that could identify patients, thereby protecting patient privacy. Common image enhancement methods were used to augment the training set and improve the robustness of model training, including rotation of plus or minus 10 degrees, random cropping, and random scaling of 80\%-120\%.

\begin{figure}[htp]
	\centering
	\includegraphics[width=1\linewidth]{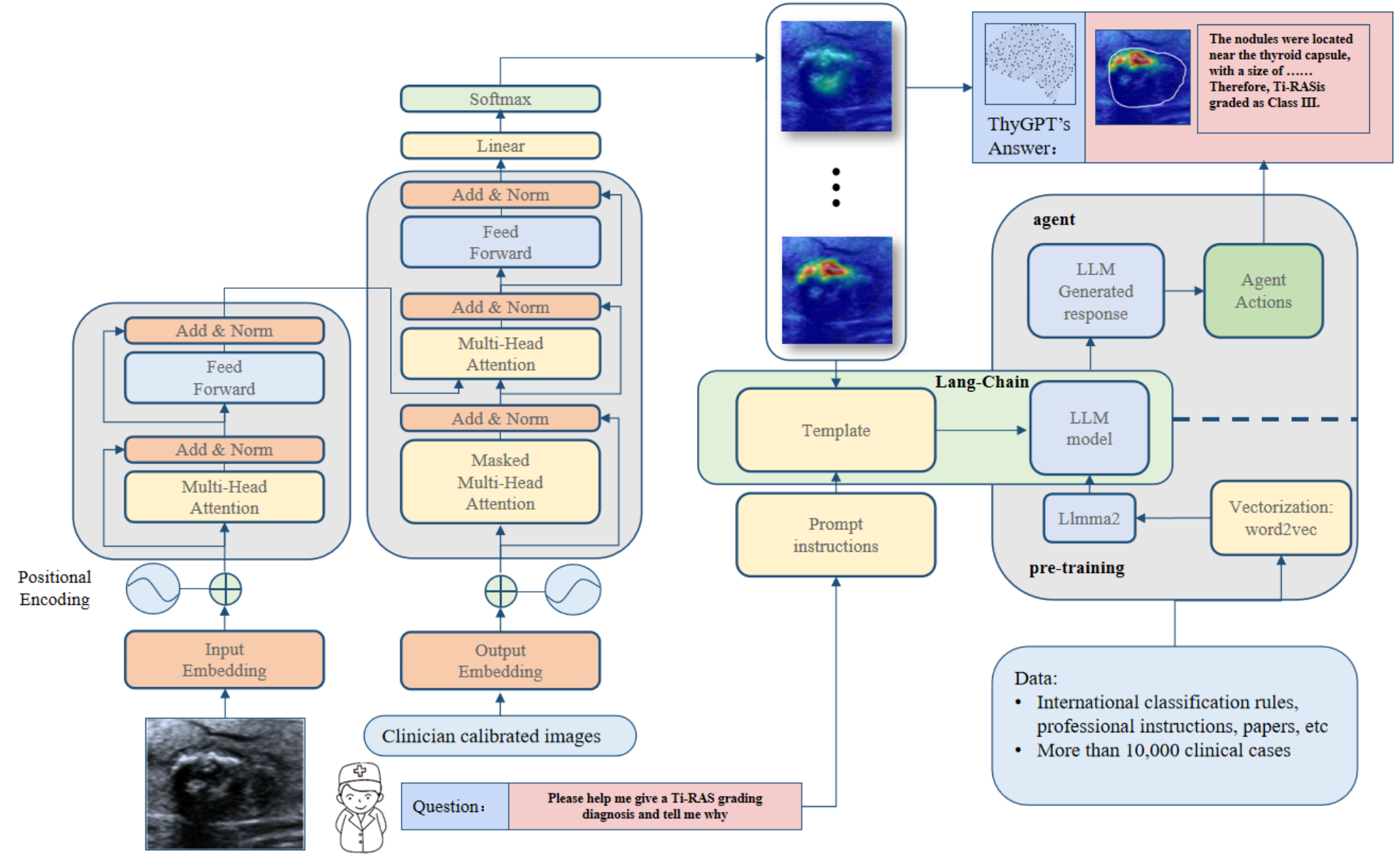}
	\caption{The internal design architecture of the proposed ThyGPT model.}
	\label{fig:cad-02}
\end{figure}

\subsection{Statistical Analysis}
The performance of the model was assessed using ROC and AUC (area under curve), sensitivity, and specificity values. In addition, we designed a comprehensive performance assessment to evaluate the ThyGPT model's effectiveness, which included the performance of radiologists reading images alone, the model's independent recognition capabilities, and the secondary judgments made by radiologists interacting with the AIGC-CAD model. The paper employed the DeLong method to calculate the 95\% confidence intervals (CIs) for the metrics. All calculations of the metrics and the plotting of the ROC curves were conducted using the Python programming language, utilizing libraries such as matplotlib and sklearn.

\section{Results}
Our experimental outcomes were validated in two independent test sets. The first independent test set comprised 1933 patients, with 947 confirmed to have malignant tumors postoperatively. The second set included 2096 patients, with 1021 confirmed to have malignant tumors postoperatively. In this study, we established a set of assistive diagnostic rules for physicians using the ThyGPT, namely: (1) initial independent assessments by both physicians and ThyGPT; (2) no modifications when agreements occur between the physician's and ThyGPT's judgments; (3) physicians lead and interrogate ThyGPT upon discrepancies in assessments; (4) ultimately, physicians may revise their initial judgments based on the responses generated by ThyGPT.
\begin{figure}[!h]
	\centering
	\includegraphics[width=1\linewidth]{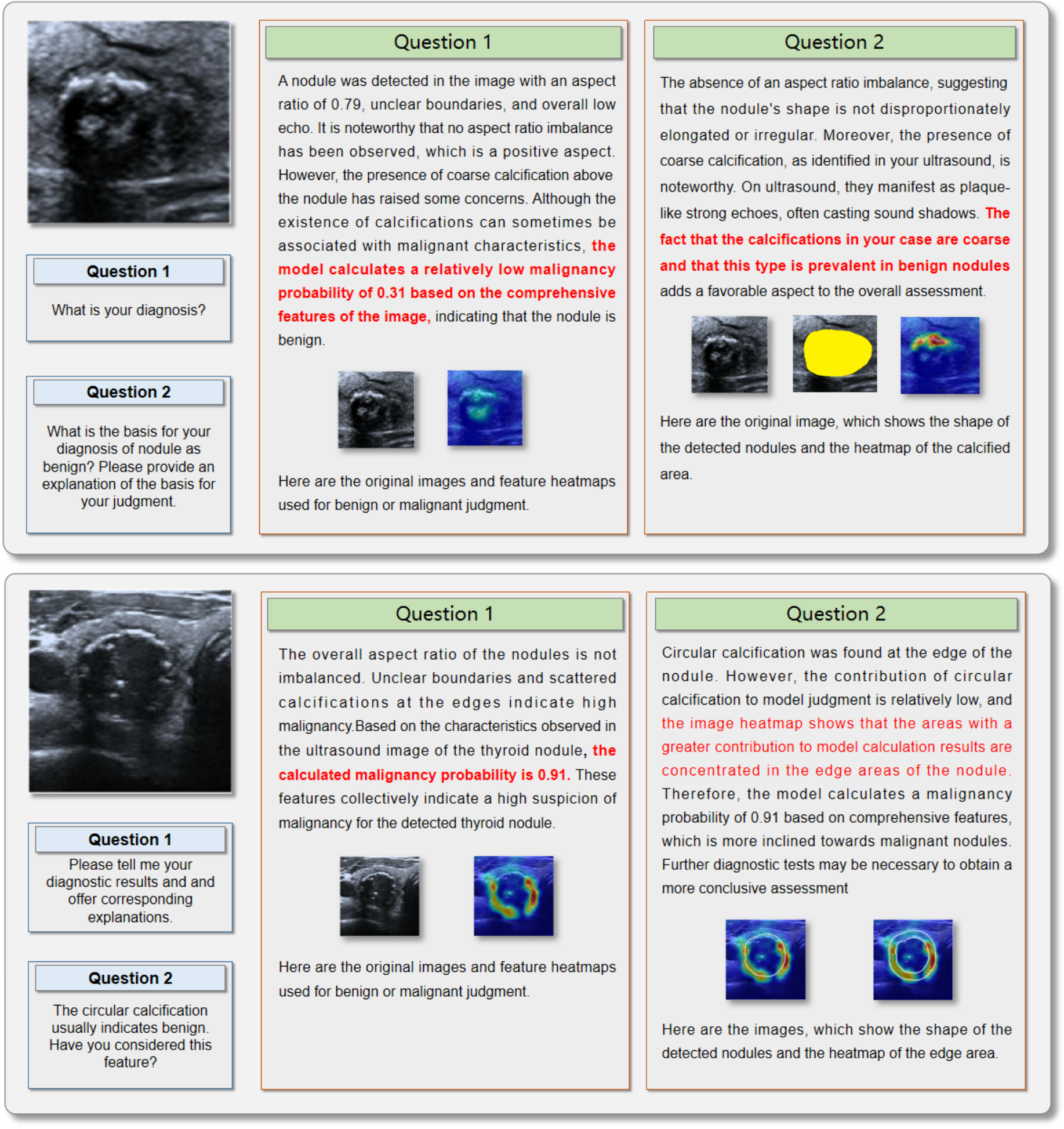}
	\caption{Sample cases with physician errors and correct ThyGPT assessments.}
	\label{fig:cad-03}
\end{figure}

Figures 3 and 4 display several representative cases. Figure 3 shows samples where clinical physicians initially erred, but after reviewing ThyGPT's analysis, they revised their diagnosis. Specifically, in Sample 1 of Figure 3, the radiology's initial diagnosis inclined towards malignancy, while ThyGPT, after scanning the image, identified the nodule as benign, providing a rationale including aspect ratio, calcification properties, and nodule composition. Consequently, the physician revised the diagnosis after considering ThyGPT's output. In the diagnosis of Sample 2, the clinician observed ring-shaped calcification, typically associated with benign nodules and thus initially identified the nodule as benign. However, ThyGPT assessed the nodule as malignant, discounting the ring-shaped calcification as a benign feature and highlighting that the key malignant features were located in the nodule's margin area. The physician accepted ThyGPT's assessment after the interaction. The detailed responses from ThyGPT are depicted in Figure 3.

Figure 4 presents cases where clinical physicians correctly diagnosed, whereas ThyGPT erred, but physicians dismissed ThyGPT's judgment after communication. For instance, in Sample 1, the pathological outcome was malignancy, but the model assessed the nodule as benign with a probability of 0.303. Upon examining the model's heat map, the physician concluded that the high-weight regions did not convey critical information and that the model's target region had diverged from the actual nodule location, rendering the model's diagnostic unreliable. Thus, the physician maintained the malignant diagnosis. For Sample 2, with a pathological result of a benign nodule, the model incorrectly assessed it as malignant. The physician deemed the heat map overly concentrated on cystic regions, lacking in reference value, questioned the model's judgment, and instructed the model to recalculate, ignoring the erroneous features in the cystic area. The model then provided a correct diagnosis following the physician's directive.

\begin{figure}[!h]
	\centering
	\includegraphics[width=1\linewidth]{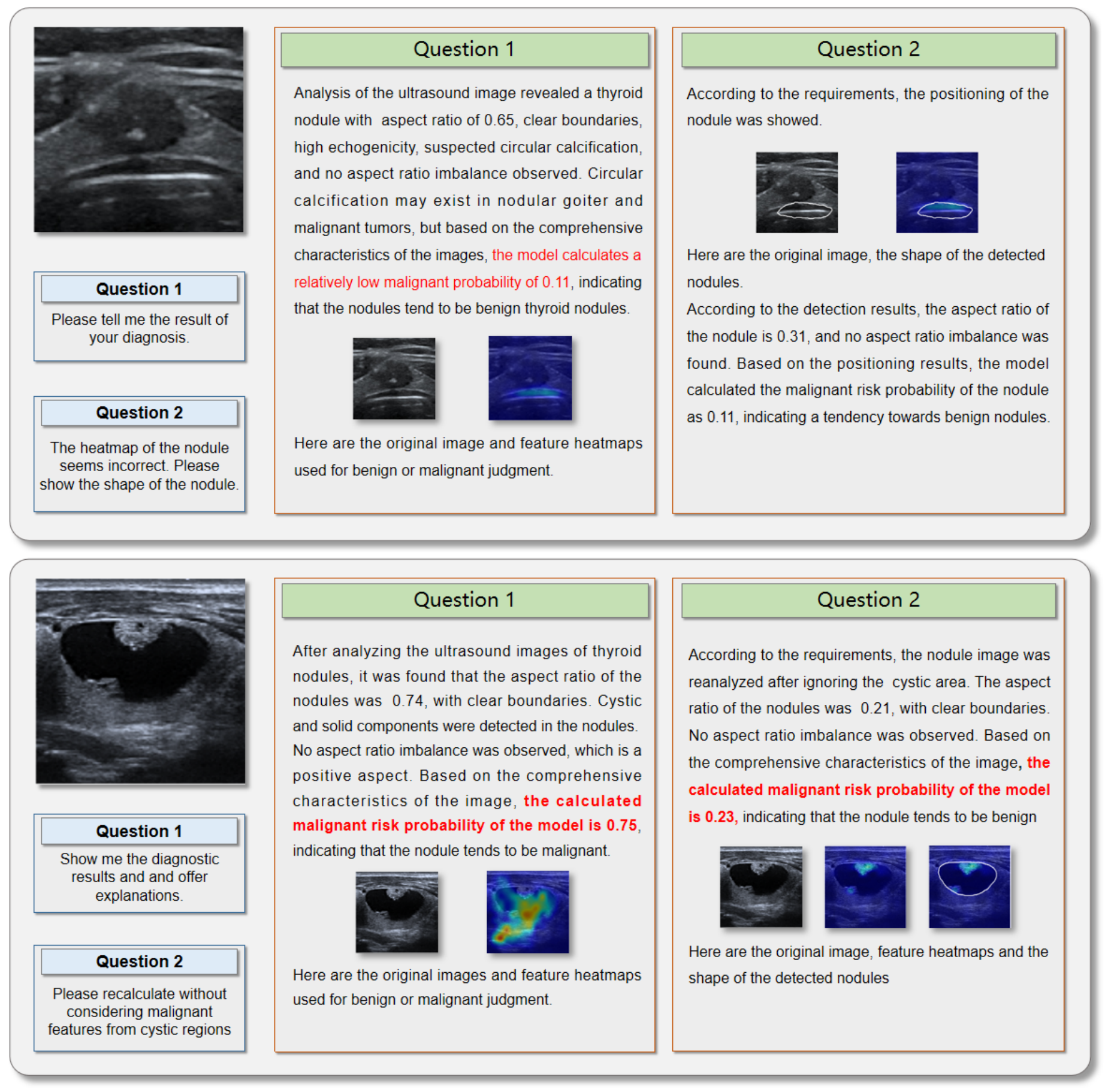}
	\caption{Sample cases with physician correctness and erroneous ThyGPT assessments.}
	\label{fig:cad-04}
\end{figure}

Figure X(a) illustrates the ROC curves of different methodologies in Test Set 1 alongside the True Positive Rate (TPR) and True Negative Rate (TNR) of four clinical physicians. Initially, we compared the performance of ThyGPT (shown in Figure X by the red curve) against several contemporary models, with ThyGPT achieving an AUC of 0.909 in Test Set 1. Additionally, we compared the outcomes of two junior physicians (5 years of experience) and two senior physicians (over 10 years of experience) both independently and with the aid of ThyGPT. Figure 6 provides a comparative analysis of sensitivity and specificity: without ThyGPT, junior physicians averaged a sensitivity of 69.8\% and 67.8\% across Test Sets 1 and 2, respectively, with specificity averaging 67.2\% and 71.7\%. When referencing traditional CAD models, junior physicians saw an increase in average sensitivity to 88.7\% and 87.1\% and specificity to 81.5\% and 85.6\% across the two test sets, respectively. Without ThyGPT, senior physicians exhibited sensitivities of 82.7\% and 82.0\% and specificities ranging from 80.5\% to 81.3\%. Upon consulting ThyGPT, senior physicians' sensitivities improved to 94.5\% and 91.1\%, with specificities rising to 89.0\% and 88.8\%. ThyGPT alone demonstrated sensitivities of 86.2\% and 84.7\% and specificities of 88.3\% and 87.0\% in the independent test sets. The data denotes that junior physicians, after consulting ThyGPT, exceeded the diagnostic performance of senior physicians, achieving a diagnostic efficacy comparable to the extensively trained ThyGPT. Moreover, senior physicians enhanced their diagnostic capabilities beyond the standalone AI model's independent judgment, reaching unprecedented levels of accuracy.

The red curve represents ThyGPT's detection curve; black crosses and X-marks represent the independent judgment curves of junior physicians; black triangles and pentagrams represent the independent judgment results of senior physicians; green crosses and X-marks indicate the diagnostic outcomes of junior physicians after referencing the explainable model; green triangles and pentagrams show the diagnostic outcomes of senior physicians after referencing the explainable model.

\begin{figure}[!h]
	\centering
	\includegraphics[width=1\linewidth]{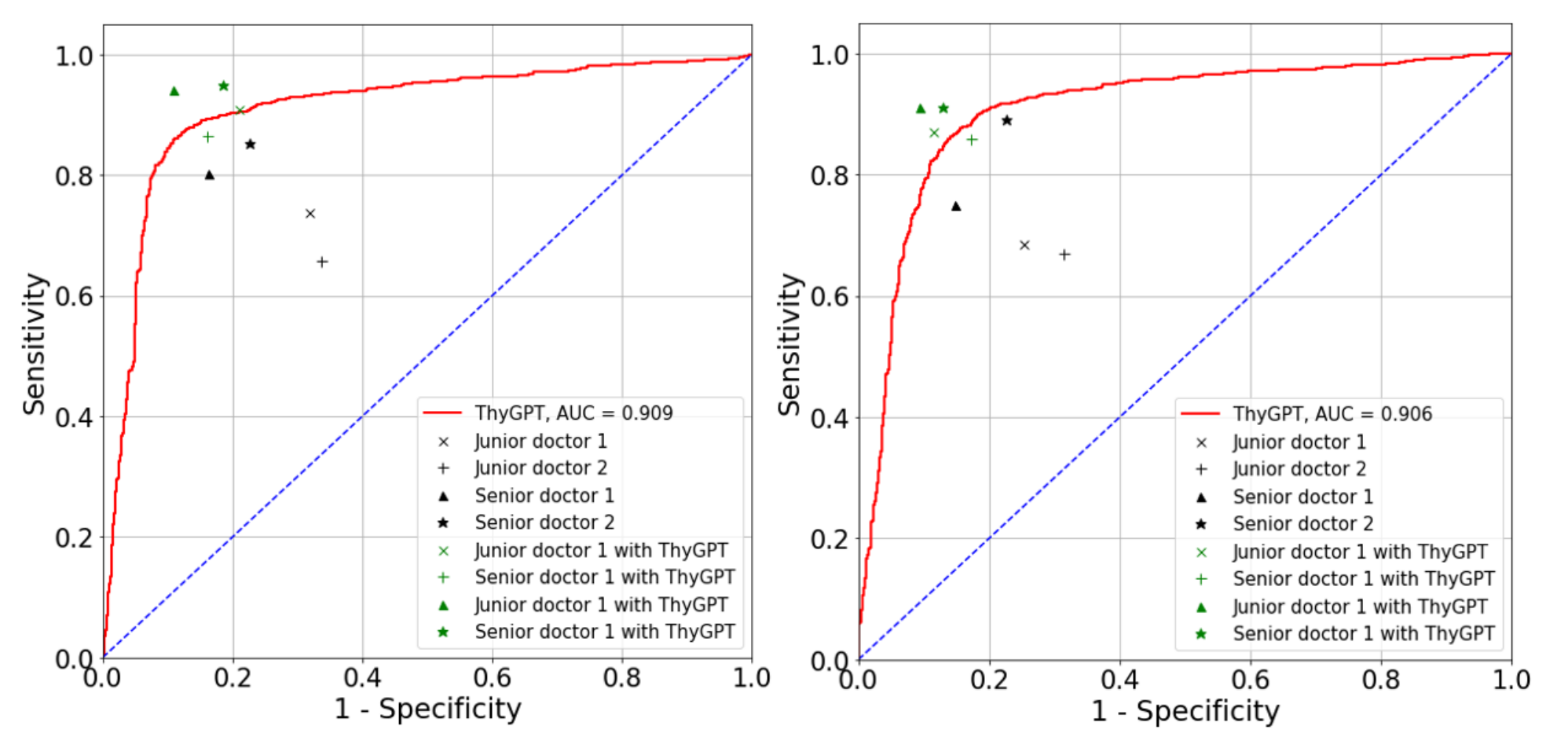}
	\caption{Clinical physicians' diagnostic results and ROC curves of deep learning methods.}
	\label{fig:cad-05}
\end{figure}

\section{Discussion}
In this study, we trained and built a large model called ThyGPT based on the next-generation generative Transformer architecture. This model is an AIGC-CAD (Artificial Intelligence-Generated Content-Computer Aided Diagnosis) system capable of in-depth interaction with clinical physicians. After training, ThyGPT successfully understood the various imaging features in thyroid nodule ultrasound images and their corresponding nodule components, displaying the contributions of each ultrasound feature of interest in the CAD model decision-making process through human-computer interaction. Our model demonstrated advantages in improving diagnostic accuracy, particularly for junior radiologists. Inter-observer variability in ultrasound image interpretation is significant, especially among junior radiologists. On two independent test sets, junior radiologists showed a 27.1\% and 28.5\% increase in sensitivity, and a 21.3\% and 19.4\% increase in specificity after referring to ThyGPT. Senior physicians experienced a 14.3\% and 11.1\% increase in sensitivity and a 10.6\% to 9.2\% increase in specificity after referencing ThyGPT. The results suggest that junior radiologists may be more receptive to ThyGPT's recommendations, potentially due to a lack of confidence and experience.

Previous studies have indicated that CAD models can outperform healthcare professionals in certain clinical outcomes. However, the reality is that traditional CAD models are not practical for standalone use in clinical practice. On one hand, traditional CAD models only output pattern recognition probabilities and cannot effectively communicate with doctors, leading to a lack of confidence in the assisted diagnosis results among doctors, patients, and healthcare institution managers. On the other hand, in actual clinical work, interferences caused by data noise, image omissions, and machine model differences far exceed the algorithm model's scope, resulting in unstable recognition probabilities. Therefore, even if CAD models are reported to have superior discriminative performance compared to doctors, real-world decisions should still be physician-led with the doctor making the final call. Compared to some previous representative CAD works, the ThyGPT model can bridge the ``interaction and understanding" gap, showing doctors, patients, and even regulators the basis for AI-assisted assessment of thyroid nodule risk. Such communication helps eliminate instability factors caused by data noise, machine model differences, or the algorithm itself, and doctors can decide whether to refer to the model's diagnostic results after observing the ThyGPT model's diagnostic rationale. To some extent, this makes AIGC-CAD systems like ThyGPT more transparent and controllable, significantly improving the confidence of clinical physicians when using CAD tools.

Our study also has some limitations: (a) Since the data comes from different machines, there are image differences between machines. Although we have used some necessary data augmentation methods to improve the model's robustness to data from different machine types, the impact of image differences on the model cannot be ignored. (b) As this study is a preliminary exploration, we only used data from a single center, even though we set up independent test sets. Further data collection for training and validation could yield a model with better robustness and stability. In the future, we will focus on the impact of the above issues and continuously refine the model to achieve optimal settings.

\bibliographystyle{unsrt}
\bibliography{references}

\end{document}